\documentclass[10pt,twocolumn,letterpaper]{article}

\usepackage[pagenumbers]{iccv} 

\definecolor{iccvblue}{rgb}{0.21,0.49,0.74}
\usepackage[pagebackref,breaklinks,colorlinks,allcolors=iccvblue]{hyperref}

\def\paperID{*****} 
\def\confName{ICCV}
\def\confYear{2025}

\title{Efficient multi-view training for 3D Gaussian Splatting}

\author{
Minhyuk Choi\textsuperscript{1}\thanks{Equal contribution} \\
\textsuperscript{1}Korea University \\
{\small \texttt{sodlqnf123@korea.ac.kr}}
\and
Injae Kim\textsuperscript{1}\footnotemark[1] \\
\textsuperscript{1}Korea University \\
{\small \texttt{dna9041@korea.ac.kr}}
\and
Hyunwoo J. Kim\thanks{Corresponding author} \\
KAIST \\
{\small \texttt{hyunwoojkim@kaist.ac.kr}}
}

\usepackage{hyperref}
\usepackage{url}
\usepackage{amsmath}
\usepackage{graphicx}
\usepackage{multirow}
\usepackage{textcomp, gensymb}
\usepackage[capitalize]{cleveref}
\usepackage{algorithm}
\usepackage{algpseudocode}
\usepackage{color}
\usepackage{colortbl}
\usepackage{amsthm}
\usepackage{booktabs}
\usepackage{subcaption}
\usepackage{makecell}
\usepackage{mathrsfs}
\newtheorem{proposition}{Proposition}

\newtheorem{lemma}{Lemma}

\newcommand{\fip}[2]{\langle #1,#2 \rangle_F}

\def \pr {\partial}
\def \bmu {\boldsymbol\mu}
\def \cl {\mathcal{L}}
\def \cw {\mathbf{w}_{\mathbf{c}}}
\def \dw {\mathbf{w}_{{\sigma}}}

\def \Ccal {\mathcal{C}}
\def \Bcal {\mathcal{B}}
\def \Scal {\mathcal{S}}

\usepackage{amssymb}
\usepackage{pifont}
\usepackage{ulem}

\newcommand{\xmark}{\ding{55}}

\usepackage{multirow} 

\begin{document}
\maketitle
\begin{abstract}
3D Gaussian Splatting (3DGS) has emerged as a preferred choice alongside Neural Radiance Fields (NeRF) in inverse rendering due to its superior rendering speed. Currently, the common approach in 3DGS is to utilize ``single-view" mini-batch training, where only one image is processed per iteration, in contrast to NeRF’s ``multi-view" mini-batch training, which leverages multiple images. We observe that such single-view training can lead to suboptimal optimization due to increased variance in mini-batch stochastic gradients, highlighting the necessity for multi-view training. However, implementing multi-view training in 3DGS poses challenges. Simply rendering multiple images per iteration incurs considerable overhead and may result in suboptimal Gaussian densification due to its reliance on single-view assumptions. To address these issues, we modify the rasterization process to minimize the overhead associated with multi-view training and propose a 3D distance-aware D-SSIM loss and multi-view adaptive density control that better suits multi-view scenarios. Our experiments demonstrate that the proposed methods significantly enhance the performance of 3DGS and its variants, freeing 3DGS from the constraints of single-view training.

\end{abstract}

\begin{figure*}[t!]
\centering
 \includegraphics[trim=0 0 0 0,clip, width=1.0\linewidth]{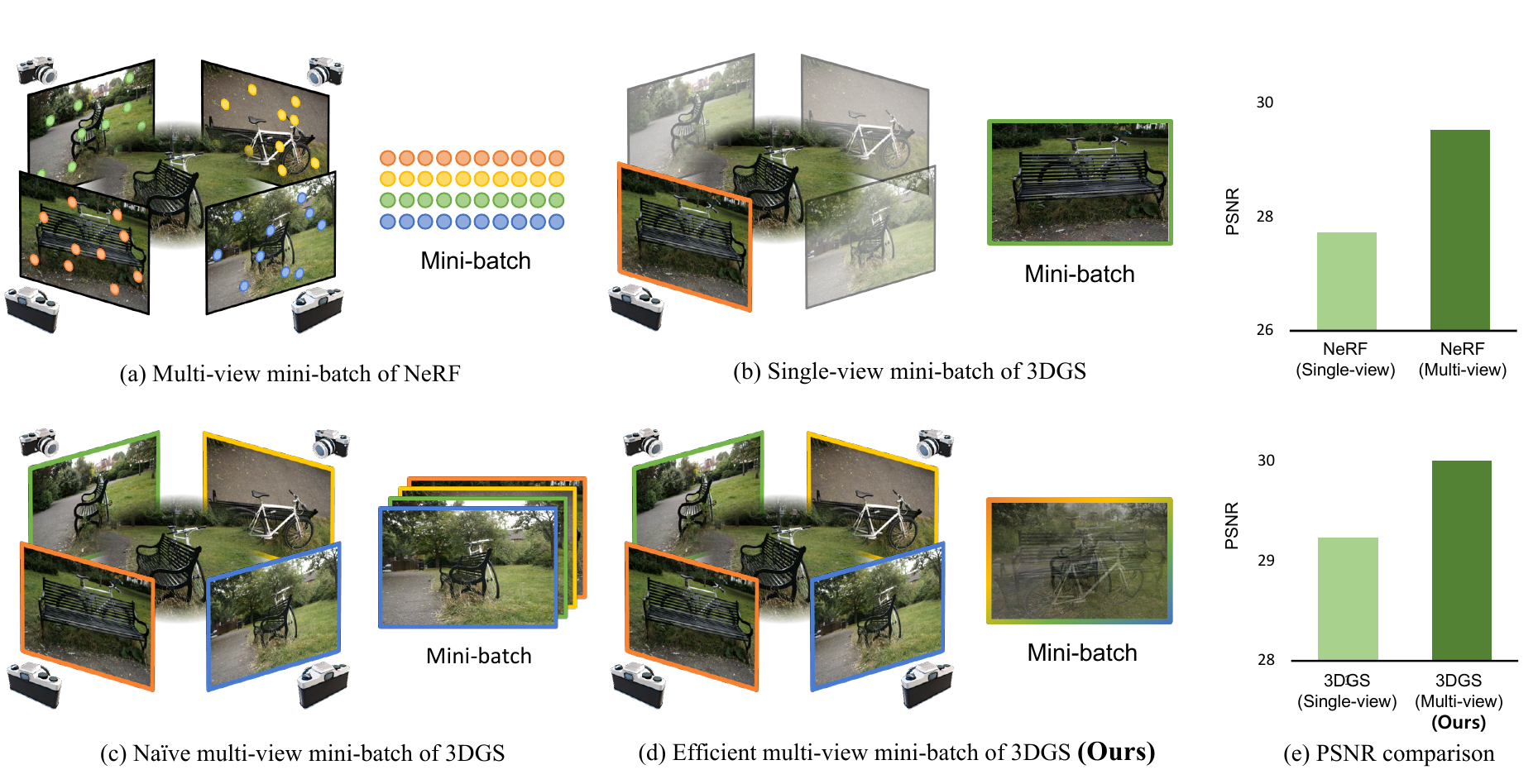}
  \caption{\textbf{Comparison between single-view and multi-view training.} We observed that single-view training degrades the efficacy of mini-batch training compared to multi-view training for both NeRF and 3DGS. However, the naïve adoption of multi-view training has limitations in terms of performance boost and considerable overhead. Our method realizes efficient multi-view training with reduced overhead, with the same number of Gaussians to the single-view 3DGS.}
 \label{fig:main}
\end{figure*}
\section{Introduction}
Deep learning-based 3D reconstructions~\citep{mildenhall2020nerf, yu2021plenoxels, mueller2022instant, kerbl3Dgaussians} are recognized as leading approaches in the inverse rendering domain. Notably, 3D Gaussian Splatting (3DGS)~\citep{kerbl3Dgaussians} has gained traction across various applications due to its exceptional rendering speed. This speed advantage arises from the differences in rendering techniques: NeRF~\citep{mildenhall2020nerf} employs ray marching, while 3DGS utilizes rasterization. Specifically, NeRF requires evaluating hundreds of points along a ray to render each pixel, whereas 3DGS splats 3D Gaussian primitives onto the image plane using an efficient CUDA-based rasterization pipeline~\citep{zwicker2001ewa}. The nature of the rasterization pipeline makes it more conducive to rendering a single image per iteration; consequently, `single-view' mini-batch training has become the de facto standard for 3DGS.
However, surprisingly, we found from a series of experiments that multi-view training leads to a smaller variance of mini-batch gradients than single-view training, leading to more stable optimization.
These observations suggest that the current single-view learning approach for 3DGS may degrade performance, while NeRF implicitly benefits from multi-view learning. \\
However, adopting multi-view training for 3DGS presents several challenges. First, due to the rasterization process, 3DGS cannot simply gather individual pixels from multiple images to form a mini-batch. Instead, it requires rendering every pixel of all images in a single iteration, which incurs considerable overhead. Second, the adaptive density control (ADC) mechanism, which modulates the number of Gaussian primitives, is based on the assumption of single-view training. Specifically, it assumes that all 2D positional gradients computed during backpropagation reside within a single plane, which is necessary for valid arithmetic operations between vectors. Without addressing this assumption, multi-view training risks performing arithmetic between vectors in different spaces, resulting in a suboptimal densification process.


In this paper, we address the challenges of efficient multi-view training in 3DGS as follows: First, we theoretically analyze the superiority of multi-view training over single-view training and provide empirical evidence of the performance limitations imposed by single-view training. Next, we propose a new rasterization method that enables 3DGS to adopt multi-view training with mitigated overhead, and we also propose a 3D distance-aware D-SSIM loss required for the new method. Lastly, we modify the densification process to eliminate the constraints of single-view training, allowing 3DGS to fully benefit from multi-view training.\\
In summary, our contributions are presented as follows.
\begin{itemize}
\item 
We provide a theoretical and empirical analysis of the efficacy of multi-view training compared to single-view training. 
\item 
We address the challenges of implementing multi-view training in 3DGS regarding additional overhead and propose solutions to mitigate these challenges. 
\item 
We eliminate the single-view assumption in the densification process, further enhancing the performance of multi-view training.
\item 
In short, our methods enable 3DGS to efficiently process multiple images simultaneously, demonstrating superior performance in extensive experiments across various datasets, which validate the effectiveness of our proposed approach.
\end{itemize}

\section{Related work}
\paragraph{Neural Radiance Fields}
Since Neural Radiance Field (NeRF)~\citep{mildenhall2020nerf} made a breakthrough in inverse rendering, NeRF has been the most favorable model for various 3D reconstruction and rendering tasks. Compared to prior works~\citep{jiang2020sdfdiff,yan2016perspective,roveri2018pointpronets} that require prior 3D representation, NeRF shows far superior results without such requirements and enables quality rendering only with images. It represents 3D space as a continuous field that outputs color and density from 3D coordinates and renders the image with volume rendering. The input and output values of the model are not restricted to the original ones. Such versatile design realizes a myriad of following works, including 4D representation~\citep{li2022neural,nerfplayer,kplanes_2023}, deformable fields~\citep{park2021nerfies,park2021hypernerf}, training with unconstrained photos~\citep{martinbrualla2020nerfw,chen2022hanerf}, editing~\citep{wang2021clip,kobayashi2022dff,instructnerf2023}, lightning~\citep{verbin2022ref,mai2023neural,liu2023nero}, generation~\citep{jain2021dreamfields} and more. While NeRF offers the advantage of good scalability, it also has the drawback of slow training and rendering speeds. Several recent works~\citep{mueller2022instant,2022tensorf,yu2021plenoxels}, including 3D Gaussian Splatting~\citep{kerbl3Dgaussians}, have sought to address these limitations of NeRF.

\paragraph{3D Gaussian Splatting}
Recently, 3D Gaussian Splatting (3DGS)~\citep{kerbl3Dgaussians} has gained recognition as a promising alternative to Neural Radiance Fields (NeRF) due to its remarkable rendering speed, which significantly outperforms that of NeRF. The fundamental difference lies in their approaches to rendering. While NeRF relies on ray marching, a technique that requires sampling thousands of points along each ray to generate a single pixel, 3DGS utilizes rasterization by splatting 3D Gaussian primitives directly onto the 2D image plane~\citep{zwicker2001ewa}. This approach bypasses the time-consuming point sampling processes of NeRF, substantially reducing rendering time and making real-time applications more feasible. The efficiency of 3DGS has made it suitable for many tasks previously dominated by NeRF. This versatility has led to its adoption across a range of domains and research areas that traditionally relied on NeRF like 3D scene generation~\citep{yi2023gaussiandreamer,abdal2024gaussian,pan2024fast,melas20243d,zhou2024gala3d,li2024dreamscene}, 4D representation~\citep{Wu_2024_CVPR,yang2023gs4d,ren2023dreamgaussian4d} and editing~\citep{silva2024contrastive, zhuang2024tip, palandra2024gsedit, wu2024gaussctrl}. As a result, there has been a noticeable shift towards 3DGS, with researchers exploring its potential in various contexts \citep{Wu_2024_CVPR,kulhanek2024wildgaussians,yang2023gs4d,tang2023dreamgaussian,jiang2023gaussianshader}.

\section{Preliminary}
\label{sec:preliminary}
\paragraph{3D Gaussian Splatting}
Given an image set with extrinsic and intrinsic camera parameters, 3DGS reconstructs the scene by optimizing 3D Gaussian primitives \{$\mathcal{G}\}$ in 3D space, each defined as mean $\boldsymbol{\mu}$, scale $S$, rotation $R$, color $\mathbf{c}$ and opacity $o$. From those parameters, we obtain a covariance matrix of Gaussians $\Sigma\!=\!RSS^T\!R^T$. Then, we project the 3D mean and covariance matrix onto the 2D plane to derive the 2D mean $\boldsymbol{\mu}'$ and covariance $\Sigma'$ for rendering~\citep{zwicker2001ewa}. To render the color $\hat{\mathbf{C}}(\mathbf{p})$ of pixel $\mathbf{p}$, 3DGS arranges and sorts $M$ Gaussians by depth that is responsible for the pixel $\mathbf{p}$ and performs alpha-blending with sorted Gaussians $\{G_{j}\}_{j=1}^M$ where $G$ is projected 2D Gaussian of 3D Gaussian $\mathcal{G}$:
\begin{equation}
\begin{split}
\hat{\mathbf{C}}(\mathbf{p}) = \sum_{i=1}^M \mathbf{c}_{i} T_{i} o_{i} G_{i}(\mathbf{p}), \;
T_{i} = \prod_{k=1}^{i-1}(1-o_{k}G_{k}(\mathbf{p})), \\
G_i(\mathbf{p}) = \exp(-\frac{1}{2}(\mathbf{p} - \boldsymbol{\mu}'_i)^T(\Sigma_i')^{-1} (\mathbf{p} - \boldsymbol{\mu}'_i)),
\end{split}
\end{equation}
where $G_i(\mathbf{p})$ is the density of $G_i$ at pixel $\mathbf{p}$.
Then, photometric $\ell_1$ loss with additional losses like D-SSIM loss can be applied to compare the rendered with the ground-truth image. The total number of 3D Gaussians varies across optimization since adaptive density control and pruning of 3DGS produce and eliminate Gaussians.


\paragraph{3DGS rasterization}
\label{paragraph:rasteriztion}
The rasterization process of 3DGS is divided into two stages: preprocessing and rendering. 
The preprocessing stage includes culling, projection, and organization of 3D Gaussian primitives for rendering. 
Then, each pixel is rendered using preprocessed Gaussians. All the steps are performed by the tile-based rasterizer. 
Specifically, the rasterizer splits the image into several tiles, and a group of threads (called \textit{block}) cooperate to preprocess Gaussians and render pixels in their assigned tiles.
A single block is allocated to a single tile, and a thread is freed when all of its peer threads in the same block have completed their work.
\begin{figure}[t]
     \centering
     \includegraphics[trim=0 10 0 -1cm,clip, width=1.0\linewidth]{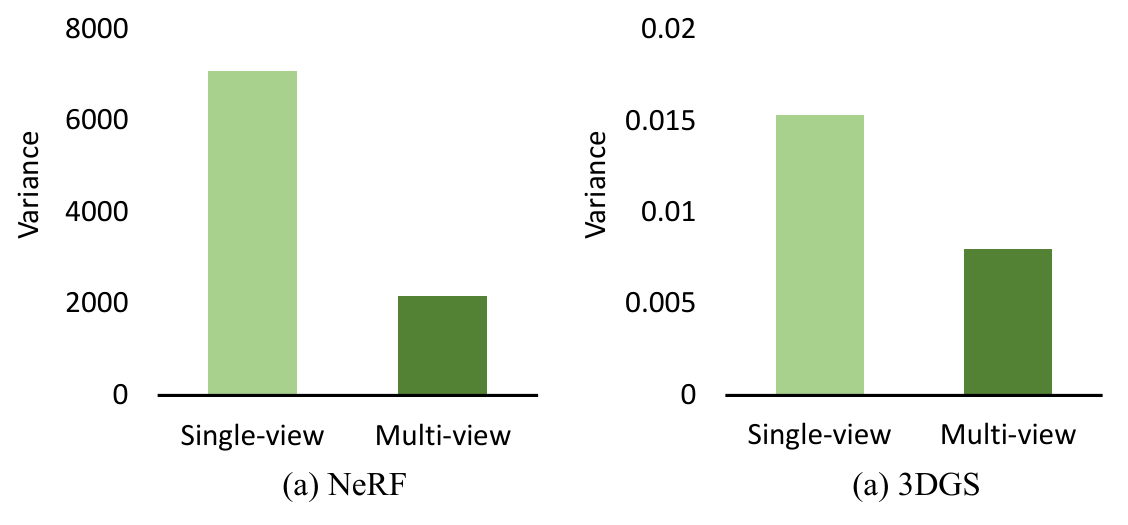}
     \caption{\textbf{Variance of mini-batch gradients.} 
     The figure indicates that multi-view mini-batch has a smaller variance of mini-batch gradients than single-view, resulting in more stable and effective optimization.
     }
     \label{fig:gradient_variance}
\end{figure}
\begin{figure*}[t!]
     \centering
     \includegraphics[trim=0 40 0 0,clip, width=1.0\linewidth]{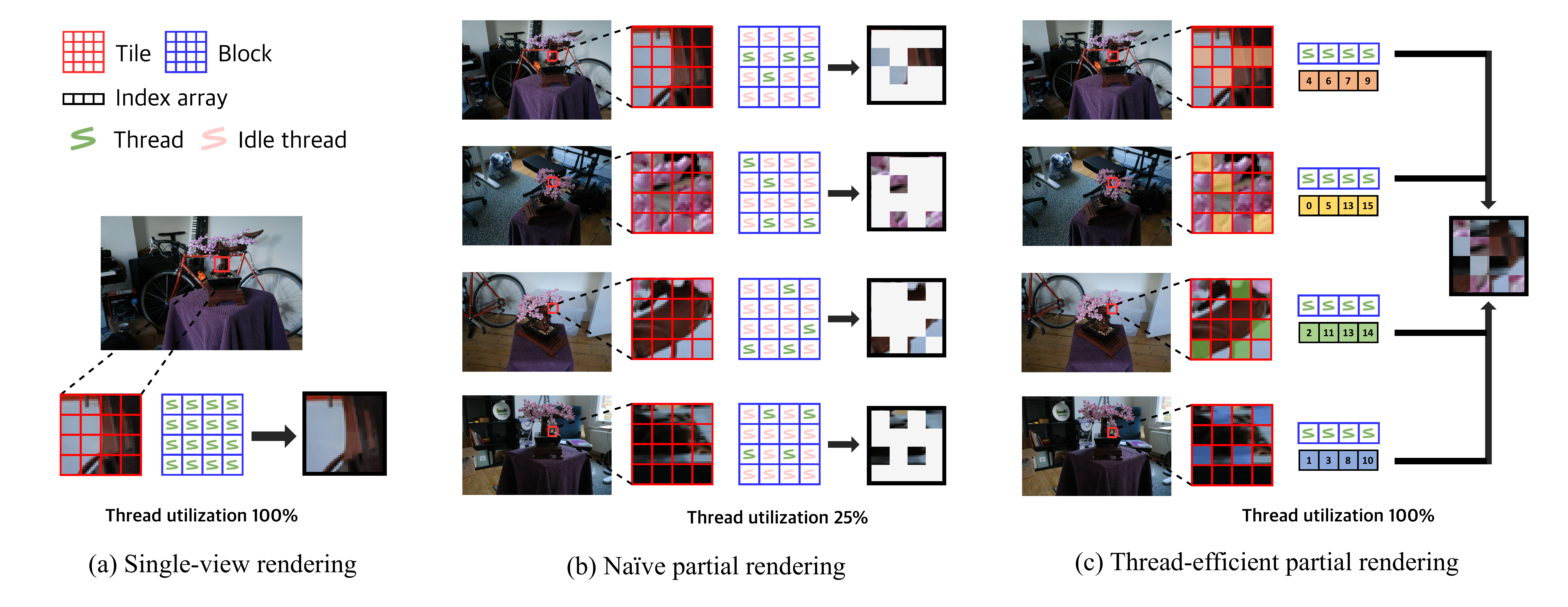}
     \vspace{-5pt}
     \caption{\textbf{Comparison of naïve partial rendering and thread-efficient partial rendering.} While the naïve implementation incurs lots of idle threads that do not participate in rendering, our thread-efficient rendering scheme does not generate such idle threads, mitigating the substantial overhead of rendering in multi-view training.}
     \label{fig:aggregated_rendering}
\end{figure*}

\section{Motivation}
\label{sec:motivation}
In this section, we present the theoretical rationale for why 3DGS should incorporate a multi-view training strategy by comparing NeRF, which trains using multi-view mini-batches. To do so, in \cref{sec:mini-batch_nerf_3dgs}, we will explain how NeRF and 3DGS define training mini-batches respectively, and in \cref{sec:mini-batch_helpful}, we will prove which one is more helpful for learning in terms of the variance of mini-batch gradients. Finally, in \cref{sec:obstacle}, we will present the challenge of multi-view training using existing rasterization methods.
\subsection{Mini-batch strategies of NeRF and 3DGS}
\label{sec:mini-batch_nerf_3dgs}
The difference in mini-batch composition between NeRF and 3DGS mainly stems from different rendering schemes in the two methods and affects the stability and efficacy of optimization.
Given $N$ images with $M$ pixels, the training data $\mathcal{D}$ consists of $R=N\!\!\times\!\!M$ rays $\mathbf{r}$ and pixel values $\mathbf{C}$, $\mathcal{D}=\{(\mathbf{r}_{i}, \mathbf{C}_{i})\}_{i\in[1,R]}$.
NeRF randomly samples data from the all training dataset $\mathcal{D}$ to form a mini-batch $\mathcal{B}$, as shown in \cref{fig:main} (a).
In contrast, 3DGS typically uses a single-view image to form a minibatch $\mathcal{B}$, as shown in \cref{fig:main} (b). We also provide a pseudocode that compares two mini-batch strategies in the supplement. 

\subsection{Efficacy of multi-view training}
\label{sec:mini-batch_helpful}
This simple difference between constructing a mini-batch as a single view or as a multi-view results in a significant difference in performance, as shown \cref{fig:main} (e); The single-view NeRF experiment sampled a mini-batch from only a single-view image, and the multi-view 3DGS experiment is the result of our proposed method.
Then, why is multi-view mini-batch more effective than single-view mini-batch?
To answer this question, we analyze mini-batch training for empirical risk minimization. 
Let $\mathbf{w}$ denote the model parameters,
then training in NeRF can be formulated as 
\begin{equation}
\min_\mathbf{w} P(\mathbf{w}) \quad P(\mathbf{w}) := \frac{1}{R}\sum_{i=1}^{R}\phi_{i}(\mathbf{w}),
\end{equation}
where $\phi_{i}(\mathbf{w}) = \cl(\mathbf{w}, r_i, \mathbf{c}_i)$ and $\cl$ is a loss function that penalizes the deviation of prediction from the label.
With mini-batch training, weights at iteration $t$ is updated with mini-batch gradient $\nabla\bar{\phi}_{k}$ as
\begin{align}
    &\mathbf{w}_{t+1} = \mathbf{w}_t - \eta_t \nabla\bar{\phi}_{k}(\mathbf{w}_t), \\
    &\text{where } \nabla\bar{\phi}_{k}(\mathbf{w}_t)=\frac{1}{|\mathcal{B}_k|}\sum_{i \in \mathcal{B}_{k}}\nabla \phi_i(\mathbf{w}_t)  \text{ and }\mathcal{B}_{k} \subset \mathcal{D}.
\end{align}
Then, it is demonstrated in \citep{zhao2014accelerating} that mini-batch training is more effective as the variance of $\nabla\bar{\phi}$, which is $\mathbb{V}_k(\nabla\bar{\phi}_k)$, decreases when we define the variance of vectors $\mathbf{x}$ as
\begin{equation}
    \label{eq:vector_variance}
    \mathbb{E}[\mathbf{x}] = \bmu, \quad 
    \mathbb{V}(\mathbf{x}) := \mathbb{E}[\Vert\mathbf{x} - \bmu\Vert^2_2].
\end{equation}
We estimate the variance above using Monte Carlo sampling to compare the variance of single-view and multi-view mini-batch gradients. 
Specifically, we freeze the model parameters, randomly sample $K$ mini-batches to compute mini-batch gradients with respect to the parameters of density MLP, and then calculate the sample variance as
\begin{equation}
\begin{split}
    \mathbb{V}_{\Bcal_k \in \mathfrak{S}}(\nabla\bar{\phi}_{k}) &= \mathbb{E}[\Vert\nabla\bar{\phi}_{k} - P(\mathbf{w}_t)\Vert^2_2] \\
    &=\mathbb{E}[\Vert\nabla\bar{\phi}_{k}\Vert_2^2] - \Vert P(\mathbf{w}_t)\Vert^2_2 \\
    &\approx \frac{1}{|\tilde{\mathfrak{S}}|}\sum_{\Bcal_k \in \tilde{\mathfrak{S}}}[\Vert\nabla\bar{\phi}_{k}\Vert^2_2] - \Big\Vert\frac{1}{|\tilde{\mathfrak{S}}|}\sum_{\Bcal_k \in \tilde{\mathfrak{S}}}[\nabla\bar{\phi}_{k}]\Big\Vert^2_2 
\end{split}
\end{equation}
where $\nabla\bar{\phi}_k=\nabla\bar{\phi}_k(\mathbf{w}_t)$, $\mathfrak{S}$ denotes a set of possible mini-batches of $\mathcal{D}$, and $\tilde{\mathfrak{S}}$ is a set of randomly sampled mini-batches.

\noindent\textit{Remarks.} As shown in~\cref{fig:gradient_variance}, the single-view mini-batch in NeRF, as well as in 3DGS, yields a larger variance of mini-batch gradients compared to the multi-view mini-batch. 
In general, a larger variance of stochastic gradient implies a larger discrepancy between the true (batch) gradient and the stochastic gradient since the expectation of stochastic gradient is equivalent to the batch gradient, \textit{i.e.}, $\mathbb{E} [\nabla\bar{\phi}_{k}] = P(\mathbf{w}_t)$ by definition.
Roughly speaking, multi-view mini-batch gradient is more accurate than single-view mini-batch.

\subsection{Obstacle to multi-view training in 3DGS}
\label{sec:obstacle}
How to apply the efficiency of multi-view mini-batch learning to 3DGS training? The simplest way is that rendering multiple images for a single iteration, as in \cref{fig:main} (c). However, it is not surprising that this implementation, although effective, incurs significant overhead, as the rasterizer has to render multiple images, which increases training time.
In conclusion, we present a novel rasterization method to address this overhead while also taking advantage of multi-view training. 

\section{Method}
We present our framework to incorporate efficient multi-view training into 3DGS. 
In \cref{sec:partial_rendering}, we propose thread-efficient partial rendering that realizes multi-view training with reasonable overhead. Then, we devise a new D-SSIM loss that can be computed in 3D space as an alternative to the current D-SSIM loss to make it compatible with our new rendering mechanism. In \cref{sec:mv_densification}, we demonstrate the current densification method of 3DGS is not compatible with multi-view training and propose a new densification strategy for multi-view training.

\subsection{Multi-view partial rendering}
\label{sec:partial_rendering}
As we mentioned in \cref{sec:obstacle}, it is obvious that rendering multiple images every iteration causes severe overhead. At least, We have to set the rasterizer to render the same amount of pixels compared to the single-view training. One solution would be \textit{partial rendering}, which makes the rasterizer only part of the pixels. Thereby, not only can the rasterizer render fewer pixels compared to naive multi-view training implementation, but also the model can be trained with the same number of pixels compared to the single-view training. However, if we adopt partial rendering, we need to deal with some problems to achieve better performance and lower overhead. We demonstrate these as follows.

\subsubsection{Thread-efficient partial rendering}
\label{sec:thread_efficient_rendering}
Even though we mentioned that partial rendering can reduce the overhead, naive implementation of partial rendering leads to inefficient parallelization due to the tiled rendering scheme of 3DGS. The problem is that GPU does not allocate threads individually; it has a minimum bundle of threads called \textit{block}. Currently, 3DGS divides the image into several \textit{tiles} whose sizes are set to be equal to the size of a block, and one block is allocated to one tile. Then, each thread in a block renders its designated pixel in a tile. If the multi-view training is simply implemented with a binary mask as in \cref{fig:aggregated_rendering} (b), threads allocated to the masked pixel must wait until all the other threads in the same block finish their job, leading to numerous idle threads and lower GPU utilization. In the extreme case, when only a single pixel is unmasked in a tile, then all the threads in a block except a single thread allocated to the unmasked pixel must wait without rendering. We remove such inefficiency as follows. Even though 3DGS aligns the size of tiles and blocks, such alignment is not mandatory. Therefore, we can decrease the block size and let multiple blocks render a single tile as in \cref{fig:aggregated_rendering} (c). Specifically, for a single tile, we sub-sample pixels to render in a tile for each viewpoint, and pass it together as an index array to the rasterizer. Then, we assign multiple blocks that have exactly the same number of threads as the number of sub-sampled pixels, with each block allocated to each viewpoint. The rendered result is similar to the merged images as depicted in \cref{fig:aggregated_rendering}. Such implementation effectively eliminates idle threads that do not participate in rendering and mitigates the overhead of multi-view training. 
\begin{figure}[t]
     \centering
     \includegraphics[trim=-30 0 -30 -10,clip, width=0.8\linewidth]{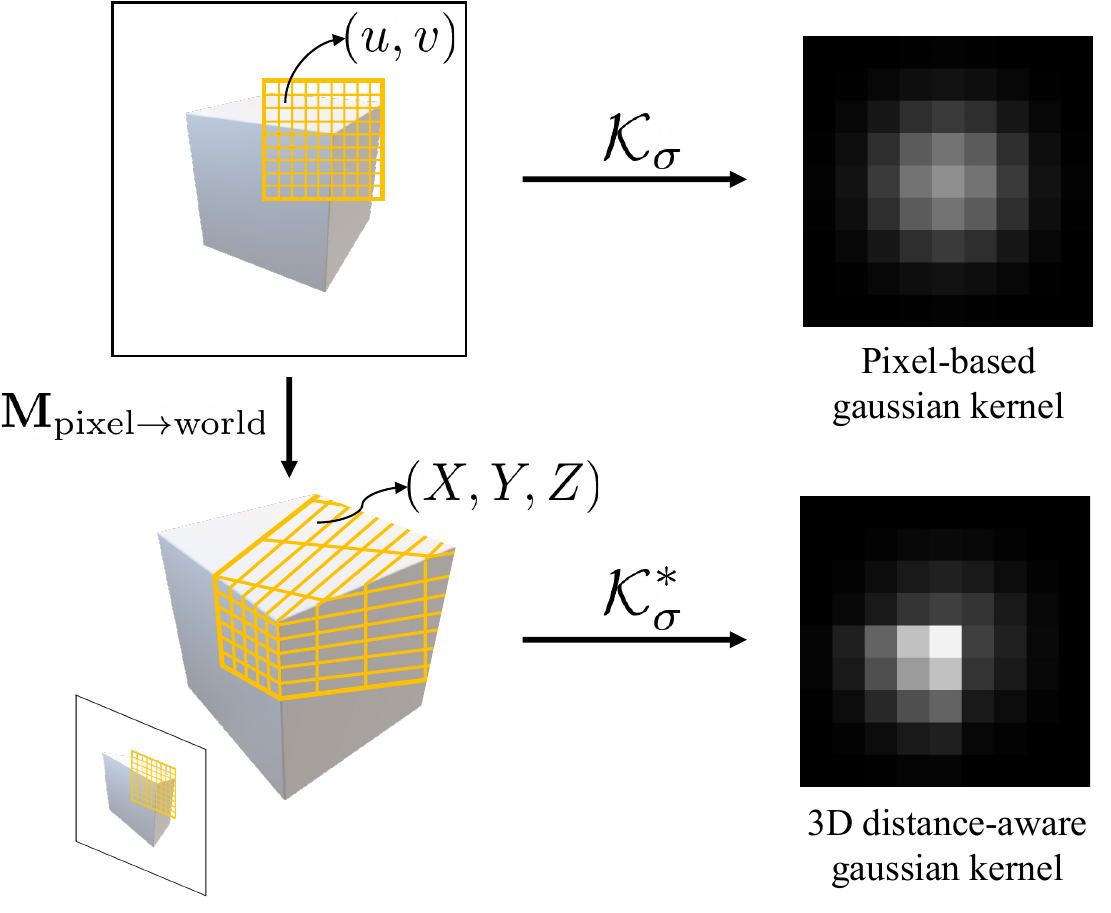}
     \caption{\textbf{3D distance-aware gaussian filter.} 
     For 3D distance-aware D-SSIM loss, we lift the points from pixel space to 3D space and compute the Gaussian kernel using the 3D distance.}
     \label{fig:3d_ssim}
\end{figure}

\subsubsection{3D distance-aware D-SSIM loss
}
3DGS uses additional D-SSIM loss for training along with the photometric $\ell_1$ loss. It penalizes low MS-SSIM (Multi-scale Structural Similarity Index Measure)~\citep{wang2004image}, which measures the structural similarity between two images by computing the mean and variance of pixel values inside a local window. While the photometric $\ell_1$ loss is vulnerable to blurry reconstruction due to its low sensitivity to outliers (fine details of the image), D-SSIM loss can produce sharper reconstruction because it prioritizes preserving local structural information, leading to reconstructions that maintain edges and textures~\citep{zhao2016loss}. 
Meanwhile, such a requirement for local information is troublesome when we apply thread-efficient partial rendering. As in \cref{fig:aggregated_rendering}, our new rendering mechanism makes pixels from multiple viewpoint lie in the sample image plane, which makes it impossible to apply conventional D-SSIM loss. Therefore, we devise 3D distance-aware D-SSIM loss that can be computed in the 3D space, such that adjacent pixels that are distant in 3D space are not forced to have structural similarity. Specifically, the original D-SSIM loss of two images is computed as follows.
\begin{align}
&\text{D-SSIM}(I_1,I_2) = 1 - \text{SSIM}(I_1, I_2)\\
&\text{SSIM}(I_1,I_2) = \frac{(2\mu_1\mu_2 + C_1)(2\tau_{12}^{ }+C_2)}{(\mu_1^2+\mu_2^2+C_1)(\tau_1^2+\tau_2^2+C_2)} 
\end{align}
where $C_1$ and $C_2$ are predefined constants, and $\mu_1$, $\mu_2$, $\tau_1^2$, $\tau_2^2$ and $\tau_{12}^{ }$ are computed with local patches $I_1$ and $I_2$.
\begin{equation}
\label{eq:ssim_mean_std}
\begin{split}
\mu_1=&\fip{I_1}{\mathcal{K}_{\sigma}}, \\
\mu_2=&\fip{I_2}{\mathcal{K}_{\sigma}}, \\
\tau_1^2=&\fip{I_1\circ I_1}{\mathcal{K}_{\sigma}} - \mu_1^2, \\ 
\tau_2^2=&\fip{I_2\circ I_2}{\mathcal{K}_{\sigma}} - \mu_2^2, \\ 
\tau_{12}^{ }=&\fip{I_1\circ I_2}{\mathcal{K}_{\sigma}} - \mu_1\mu_2,
\end{split}
\end{equation}
where $\mathcal{K}_{\sigma}$ is 2D Gaussian kernel, $\fip{I}{\mathcal{K}_{\sigma}} $ is Frobenius inner product, and $\circ$ is element wise product.
\begin{align}
    \mathcal{K}_{\sigma}(u,v)=\frac{1}{2\pi\sigma^2}\exp\Big(-\frac{u^2+v^2}{2\sigma^2}\Big),
\end{align}
where $(u,v)$ is a pixel coordinate where the center of the kernel is defined as the origin. To obtain 3D distance-aware D-SSIM loss, we utilize a 3D distance-aware Gaussian kernel $\mathcal{K}_{\sigma}^*$, which is shown in \cref{fig:3d_ssim}. 3D surface points corresponding to each pixel are calculated by using a matrix $\mathbf{M}_{\text{pixel}\rightarrow\text{world}}$ which transforms the point from the pixel space to the world space, where the coordinate corresponding to the center of the kernel is defined as the origin. $\mathcal{K}_{\sigma}^*$ is formulated as
\begin{equation}
\begin{split}
    \mathcal{K}^*_{\sigma}(u, v)=\frac{1}{2\pi\sigma^2}\exp(-\frac{X^2+Y^2+Z^2}{2\sigma^2}), \\
    \text{where} \ \ (X, Y, Z, 1)^T=\mathbf{M}_{\text{pixel}\rightarrow\text{world}}(u, v, 1)^T.
\end{split}
\end{equation}
A matrix $\mathbf{M}_{\text{pixel}\rightarrow\text{world}}$ can be obtained using the predicted depth from the rasterizer. Finally, we obtain the 3D distance-aware D-SSIM loss by replacing $\mathcal{K}_{\sigma}$ in \cref{eq:ssim_mean_std} with $\mathcal{K}_{\sigma}^*$. 
By utilizing the D-SSIM loss in 3D space, the issue of two adjacent pixels originating from different viewpoints is mitigated, as they have a low weight for each other.

\subsection{Densification in multi-view training}
\label{sec:mv_densification}
\subsubsection{Incompatability of ADC to multi-view training}
\begin{figure}[t]
     \centering
     \includegraphics[trim=0 0 0 0,clip, width=0.85\linewidth]{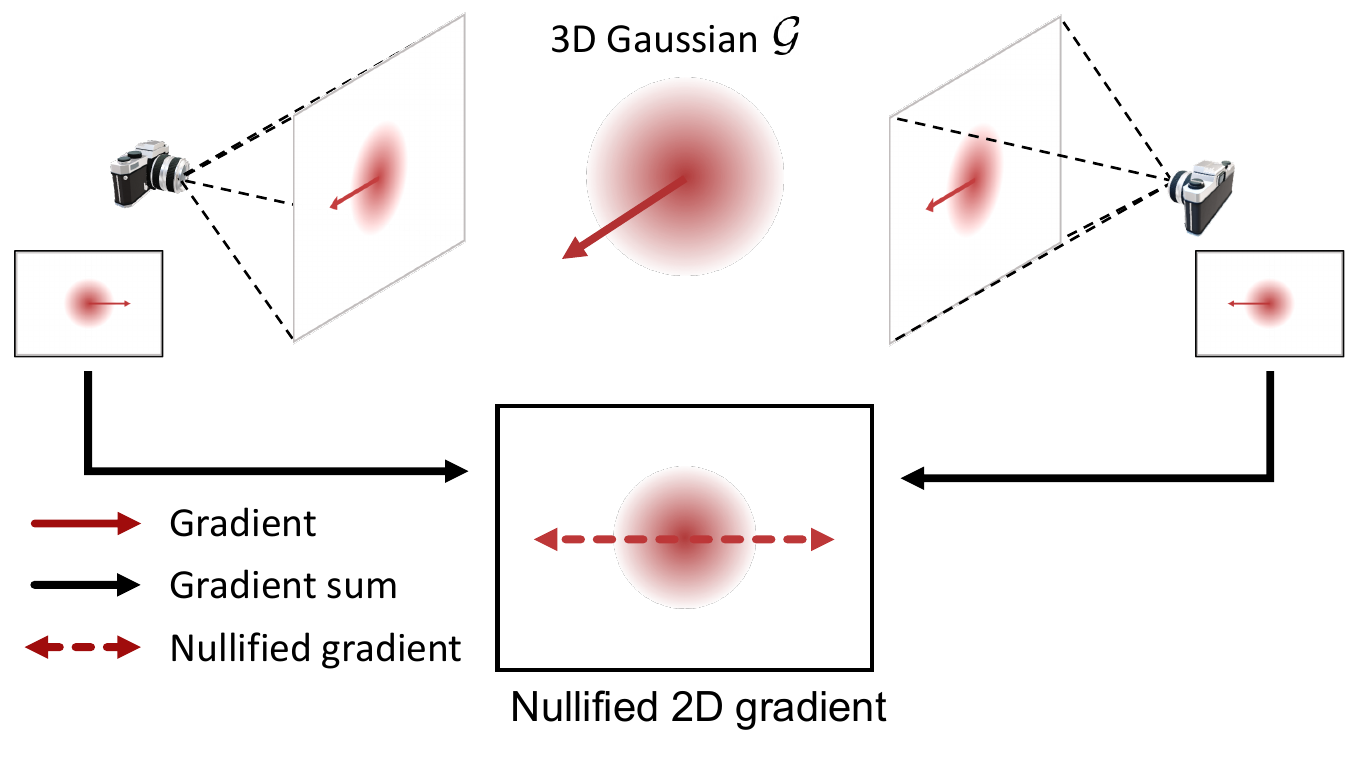}
     
     \vspace{-5pt}
     \caption{\textbf{Invalid arithmetic of gradients in different spaces.} Even though both 2D positional gradients derive the same 3D gradients, the addition of those vectors results in the zero vector. Likewise, adding 2D positional gradients in different spaces is invalid, leading to unintended consequences.}
     \label{fig:gradient}
\end{figure}

3DGS helps Gaussian primitives that struggle with reconstructing high-frequency regions through the densification process called adaptive density control (ADC). To figure out such Gaussians, ADC utilizes view-space 2D positional gradients to sort out and densify Gaussians with large errors. 2D positional gradients of each Gaussian primitive are computed by adding 2D positional gradients from pixels that each Gaussian covers, and the norm of those gradients is accumulated for multiple iterations. For every predefined interval, Gaussians with a larger mean than a predefined threshold are chosen, and ADC splits large Gaussians to have smaller sizes and clones small Gaussians. \\
Meanwhile, we can not apply ADC directly to multi-view training. As we mentioned, ADC involves the addition of 2D positional gradients, namely vectors. And vectors in different spaces can not be added. In multi-view training, each Gaussian can be projected onto image planes of multiple viewpoints, which can lead a Gaussian to receive gradients from pixels in different image planes. We provide an example situation where this invalid arithmetic operation causes a problem in \cref{fig:gradient}, which depicts an extreme case where two 2D positional gradients are nullified even though two gradients represent the same 3D space gradient.

     
\subsubsection{Multi-view Adaptive density control}
We discussed the problem of ADC in multi-view training: It causes invalid addition of gradients in different view spaces. We propose the following method to resolve this problem.
\paragraph{Space-independent error metric}
The core of densification is that Gaussians with a large error deserve densification, and the design of such an error is entirely our choice. ADC currently computes the error metrics for each Gaussian as
\begin{equation}
E_{\text{old}}(\mathcal{G})=\Big\Vert\sum_{k=1}^N \sum_{V_{\mathbf{p}_i} = k}\nabla_{\mathbf{p}_i} \mathcal{L}\Big\Vert_2,
\end{equation}
where $\{\mathbf{p}_i\}$ are the pixels a Gaussian cover, and each pixel $p_i$ comes from a viewpoint $V_{\mathbf{p}_i}$ among $N$ viewpoints.\\
Then, we can remove the invalid vector arithmetic by changing \textit{add and norm} to \textit{norm and add}. Then, we devise two different ways to compute the error metric as follows
\begin{align}
&E_1(\mathcal{G}) = \sum_{k=1}^N \sum_{V_{\mathbf{p}_i} = k}\Vert\nabla_{\mathbf{p}_i}\mathcal{L}\Vert_2, \\
&E_2(\mathcal{G}) = \sum_{k=1}^N \Big\Vert\sum_{V_{\mathbf{p}_i} = k}\nabla_{\mathbf{p}_i}\mathcal{L}\Big\Vert_2.
\end{align}
Compared to adding vectors, adding \textit{norm} remains valid even though vectors are placed in different spaces. 
We empirically found that $E_1$ is good at splitting, and $E_2$ is good at cloning.

\begin{table*}[t]
    \centering
    \resizebox{0.9\linewidth}{!}{
    \setlength{\tabcolsep}{4pt}
    \begin{tabular}{@{}c l cccc@{}} 
    \toprule
    && MipNeRF-360~\citep{barron2022mipnerf360} & OMMO~\citep{lu2023largescaleoutdoormultimodaldataset} & Deep Blending~\citep{DeepBlending2018} & Tank \& Temples ~\citep{Knapitsch2017} \\
    && \footnotesize{PSNR$\uparrow$ / SSIM$\uparrow$ / LPIPS$\downarrow$} & \footnotesize{PSNR$\uparrow$ / SSIM$\uparrow$ / LPIPS$\downarrow$} & \footnotesize{PSNR$\uparrow$ / SSIM$\uparrow$ / LPIPS$\downarrow$} & \footnotesize{PSNR$\uparrow$ / SSIM$\uparrow$ / LPIPS$\downarrow$}  \\
    \midrule 
    
    & NeRF~\cite{mueller2022instant}            & 24.85 / 0.659 / 0.426 & - & -                     & -                     \\
    & Plenoxels~\cite{yu2021plenoxels}          & 23.63 / 0.670 / 0.443 & - & 23.06 / 0.795 / 0.510 & 21.08 / 0.710 / 0.379 \\
    & INGP-Big~\cite{mueller2022instant}        & 26.75 / 0.751 / 0.299 & - & 24.96 / 0.817 / 0.390 & 21.92 / 0.745 / 0.305 \\
    & MipNeRF~\cite{barron2021mipnerf}          & 27.60 / 0.806 / 0.251 & - & -                     & -                     \\
    & MipNeRF360~\cite{barron2022mipnerf360}    & 29.23 / 0.845 / 0.207 & - & 29.40 / 0.902 / 0.245 & 22.22 / 0.759 / 0.257 \\
     
    \midrule
    & 3DGS~\citep{kerbl3Dgaussians} & 29.28 / 0.878 / 0.167 & 29.24 / 0.901 / 0.162 & 29.55 / 0.902 / 0.244 & 23.69 / 0.845 / 0.177 \\
    & + Ours & \textbf{29.74} / \textbf{0.886} / \textbf{0.154} & \textbf{30.01} / \textbf{0.911} / \textbf{0.143} & \textbf{29.64} / \textbf{0.903} / \textbf{0.238} & \textbf{24.02} / \textbf{0.856} / \textbf{0.158} \\
    \midrule
    & 3DGS-MCMC~\citep{kheradmand20243d} & 29.83 / 0.895 / 0.145 & 29.81 / 0.914 / 0.140 & 29.57 / 0.905 / 0.240 & 24.59 / 0.869 / 0.145 \\
    & + Ours$^\dag$ & \textbf{30.42} / \textbf{0.901} / \textbf{0.138} & \textbf{30.85} / \textbf{0.923} / \textbf{0.128} & \textbf{30.11} / \textbf{0.914} / \textbf{0.224} & \textbf{25.06} / \textbf{0.876} / \textbf{0.144} \\
    \bottomrule
    \end{tabular}
    }
    \caption{
        \textbf{Quantitative results of baselines and our model.} Our method improves the performance of not only 3DGS~\cite{kerbl3Dgaussians}, but also its state-of-the-art variant 3DGS-MCMC~\cite{kheradmand20243d} even further. Ours$^\dag$ means that only the multi-view training strategy is applied.
    }
    \label{tab:main}
\end{table*}

\section{Experiments}
\begin{figure*}[t]
\centering
\includegraphics[width=0.9\textwidth, trim =5 0 5 0, clip]{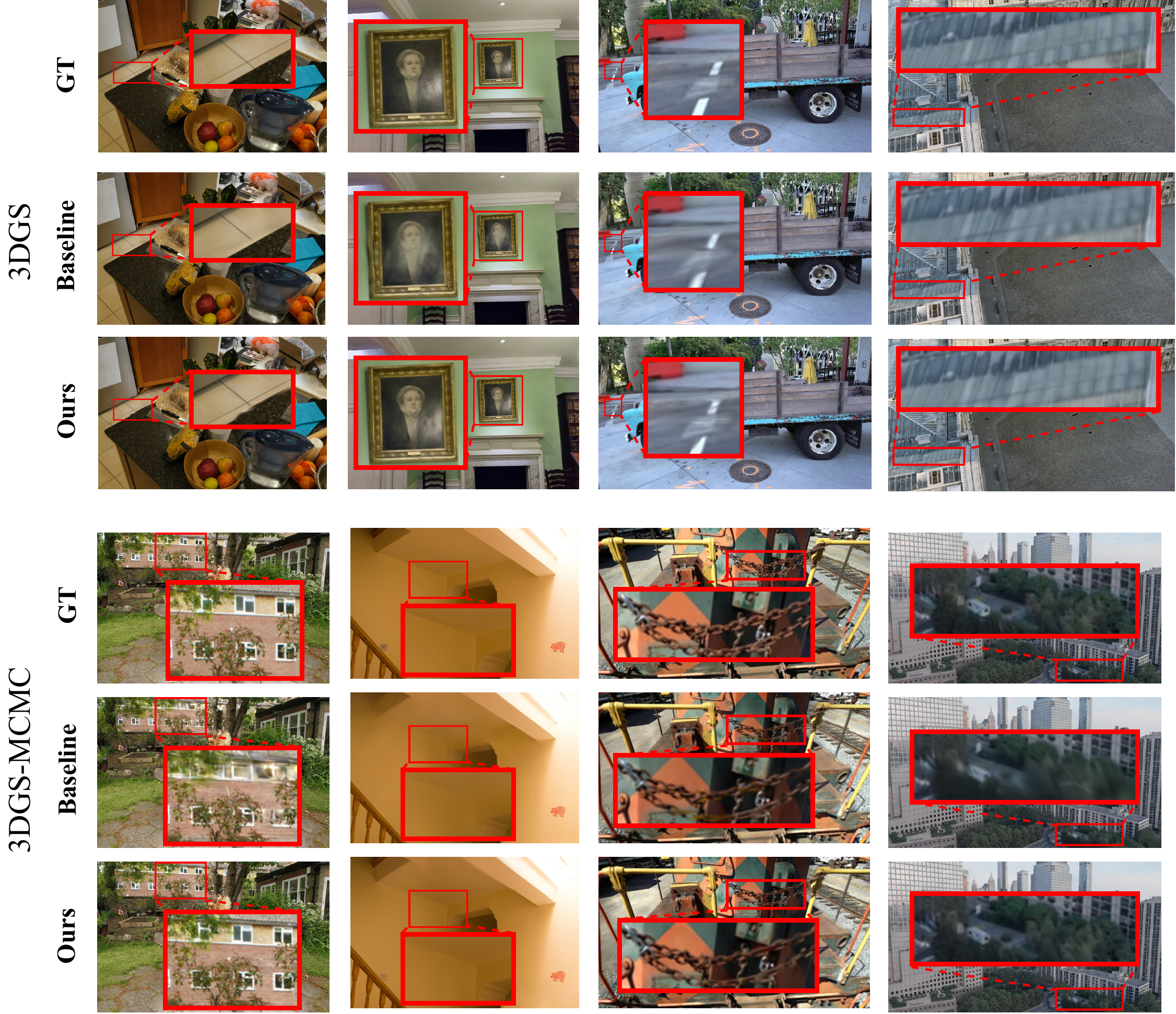}
\caption{\textbf{Qualitative results of baselines and our model.} We provide comparison of 3DGS~\citep{kerbl3Dgaussians} and 3DGS-MCMC~\citep{kheradmand20243d}. Our method successfully enhances both models, proving its compatibility with a variety of models.}
\label{fig:qualitative}
\end{figure*}

We add our methods to two baselines, original 3DGS~\citep{kerbl3Dgaussians} and 3DGS-MCMC~\citep{kheradmand20243d}, with the number of images used for each iteration fixed to 4 across experiments. We do not apply methods in \cref{sec:mv_densification} for experiments on 3DGS-MCMC because they do not use an adaptive density control process. For a fair comparison, the number of Gaussians in all experiments does not exceed the number provided by 3D-MCMC, and the average of three repeated experiments is reported. All experiments were performed on a GeForce RTX 3090 GPU.
\paragraph{Datasets and metrics}
Datasets include Deep blending~\citep{DeepBlending2018}, Tanks and Temples~\citep{Knapitsch2017}, MipNeRF-360~\citep{barron2022mipnerf360} and OMMO~\citep{lu2023largescaleoutdoormultimodaldataset}. Because 3DGS does not support the radial camera in OMMO dataset, we apply undistortion to OMMO dataset using COLMAP~\cite{schonberger2016structure}. We use three standard metrics to quantitatively assess the performance: PSNR, SSIM~\citep{wang2004image}, and LPIPS~\citep{zhang2018unreasonable}. We conduct all the experiments three times and report their average to compensate for the randomness of 3DGS. 
\paragraph{Quantitative and qualitative comparison}
\cref{tab:main} shows the results of novel view synthesis over four datasets. For a fair comparison, we set the final number of Gaussians to be the same as the number used in 3DGS-MCMC, which is the same number of Gaussians reached when training the basic 3DGS. It verifies that our approach consistently improves the results over the original 3DGS~\citep{kerbl3Dgaussians} and its state-of-the-art variant 3DGS-MCMC~\citep{kheradmand20243d}. In particular, the performance improvement of the 3DGS-MCMC proves that multi-view training itself is effective. It also shows that our method is not limited to just the original 3DGS. We also provide the qualitative novel-view synthesis results in \cref{fig:qualitative}. We highlight the differences within the images with a red box. It is observed that the models trained with the multi-view training can learn finer details.

\begin{figure}
\begin{minipage}{1.0\linewidth}
        \vspace{1mm}
  \centering\resizebox{1.0\linewidth}{!}{
    \begin{tabular}{ccccccc} 
        \toprule
            \multirow{2}{*}{Loss type}& \multicolumn{3}{c}{\small3DGS~\citep{kerbl3Dgaussians}} & \multicolumn{3}{c}{\small3DGS-MCMC~\citep{kheradmand20243d}} \\
        \cmidrule(lr){2-4} \cmidrule(lr){5-7}
        & \small PSNR $\uparrow$ & \small SSIM $\uparrow$ & \small LPIPS $\downarrow$ & \small PSNR $\uparrow$ & \small SSIM $\uparrow$ & \small LPIPS $\downarrow$ \\ 
        \midrule
        only $\ell_1$& 29.19 & 0.845 & 0.219 & 29.65 & 0.877 & 0.149   \\
        only $\ell_2$& 27.79 & 0.770 & 0.323 & 29.46 & 0.861 & 0.211   \\
        $\ell_1$ + $\ell_\text{D-SSIM}$ & 29.55 & 0.865 & 0.192 & 30.02 & 0.895 & 0.146   \\
        $\ell_1$ + $\ell_\text{3D D-SSIM}$ & \textbf{29.74} & \textbf{0.886} & \textbf{0.154} & \textbf{30.42} & \textbf{0.901} & \textbf{0.138}  \\
        
    \bottomrule
    \end{tabular}
    }
  \captionof{table}{We verify the efficacy of our masked D-SSIM loss for multi-view training. Results are the average values on the MipNeRF-360 dataset. }
  \label{tab:ablation_ssim}
\end{minipage}
\end{figure}

\begin{table}
\begin{minipage}{1.0\linewidth}
  \centering\resizebox{0.8\linewidth}{!}{
    \begin{tabular}{cccc} 
        \toprule
        Multi-view ADC & \small PSNR $\uparrow$ & \small SSIM $\uparrow$ & \small LPIPS $\downarrow$ \\ 
        \midrule
        \xmark & 29.56 & 0.880 & 0.174 \\
        \checkmark & \textbf{29.74} & \textbf{0.886} & \textbf{0.154}  \\ 
    \bottomrule
    \end{tabular}
    }
  \captionof{table}{We report the performance of 3DGS multi-view mini-batch training with and without our multi-view ADC. Results are the average values on the MipNeRF-360 dataset.}
  \label{tab:ablation_adc}
\end{minipage}
\end{table}

\begin{figure}
\begin{minipage}{1.0\linewidth}
  \centering\resizebox{0.9\linewidth}{!}{
    \begin{tabular}{lcccc}
        \toprule
        {3DGS~\citep{kerbl3Dgaussians}}\,\,\,\, & \small Time $\downarrow$ & \small PSNR $\uparrow$ & \small SSIM $\uparrow$ & \small LPIPS $\downarrow$ \\ 
        \midrule
        Full  & 127m  & 29.74 & 0.882 & \textbf{0.154}  \\
        Partial & 105m  & 29.71 & 0.883 & 0.156  \\
        Ours & \textbf{50m}  & \textbf{29.74} & \textbf{0.886} & \textbf{0.154}  \\
        \bottomrule
    \end{tabular}
    }

  \centering\resizebox{0.9\linewidth}{!}{
    \begin{tabular}{lcccc}
        \toprule
        {MCMC~\citep{kheradmand20243d}} & \small Time $\downarrow$ & \small PSNR $\uparrow$ & \small SSIM $\uparrow$ & \small LPIPS $\downarrow$ \\ 
        \midrule
        Full  & 181m  & 30.34 & 0.898 & \textbf{0.138}  \\
        Partial & 135m  & 30.36 & 0.899 & 0.140  \\
        Ours & \textbf{74m}  & \textbf{30.42} & \textbf{0.901} & \textbf{0.138} \\
        \bottomrule
    \end{tabular}
    }

  \captionof{table}{We compare the performance and overhead of our thread-efficient rendering with two naïvely implemented multi-view mini-batch training methods. Results are the average values on MipNeRF-360 dataset. MCMC denotes for 3DGS-MCMC.}
  \label{tab:ablation_ag}
\end{minipage}
\end{figure}

\begin{figure}
\begin{minipage}{1.0\linewidth}
  \centering\resizebox{1.0\linewidth}{!}{
    \begin{tabular}{llcccc}
        \toprule
        & & \small PSNR $\uparrow$ & \small SSIM $\uparrow$ & \small LPIPS $\downarrow$ & \small Time $\downarrow$ \\ 
        \midrule
        \multirow{2}{*}{3DGS}  & Base  & 29.28 & 0.878 & 0.167 & \textbf{26m}  \\
        & Ours$^*$ & \textbf{29.43} & \textbf{0.884} & \textbf{0.157} & \textbf{26m}  \\
        \midrule
        \multirow{2}{*}{3DGS-MCMC}  & Base  & 29.83 & 0.895 & 0.145 & 46m  \\
        & Ours$^*$ & \textbf{30.25} & \textbf{0.900} & \textbf{0.142} & \textbf{37m}  \\
    \bottomrule
    \end{tabular}
    }
  \captionof{table}{\textbf{Comparison with similar training time.} By halving the training iterations and the iteration-related hyperparameters, we make the training time comparable to that of the baselines. The results are averaged over the MipNeRF-360 dataset.}
  \label{tab:ablation_sametime}
\end{minipage}
\end{figure}

\paragraph{Ablation \& Additional experiments}
We verify whether our methods are really helpful for multi-view training. To evaluate our 3D distance-aware D-SSIM loss, we set as baselines a model that is optimized without D-SSIM loss at all and a model that uses conventional D-SSIM loss, with pixels of each viewpoint separated to compute D-SSIM loss for each viewpoint. In addition, we try optimization with $\ell_2$ loss without D-SSIM loss, which is known to be more sensitive to outliers and has been a standard choice for NeRF. As shown in \cref{tab:ablation_ssim}, conventional D-SSIM loss is not compatible with partial rendering, even though we separate pixels of each viewpoint. Furthermore, higher sensitivity to outliers of $\ell_2$ loss is still not enough to be comparable with D-SSIM loss, especially in SSIM and LPIPS metrics. Next, we evaluate our modified densification process for multi-view training strategies in \cref{tab:ablation_adc}. It shows that our modified multi-view adaptive density control further improves the multi-view training performance by rectifying the single-view training assumption of ADC. Finally, we evaluate our thread-efficient partial rendering in \cref{tab:ablation_ag}. For comparison, we set two naïvely implemented multi-view training as baselines. First, we simply render the full image multiple times. Second, we implement partial rendering by passing a binary mask for each viewpoint in a mini-batch to the rasterizer. \cref{tab:ablation_ag} shows that our thread-efficient partial rendering successfully reduces the overhead of multi-view training without any significant performance drop. We have an additional experiment to assure that improved performance does not mainly stem from longer training time. We halve the training iterations and iteration-related hyperparameters in \cref{tab:ablation_sametime}. It clearly shows that even for the similar, or even shorter training time, our method shows superior results.


\section{Conclusion}
This paper introduces a novel, efficient multi-view training method for 3D Gaussian Splatting (3DGS), overcoming the limitations of previous single-view approaches. We provide theoretical and empirical evidence supporting the benefits of multi-view training and then propose a novel partial rendering technique to reduce computational overhead, a new 3D distance-aware D-SSIM loss function, and a modified density control mechanism. Experiments on various datasets demonstrate significant performance improvements over 3DGS and 3DGS-MCMC. The most important point is that our method is readily adaptable to enhance the performance of any 3DGS method.
{
    \small
    \bibliographystyle{ieeenat_fullname}
    \bibliography{main}
}
\newpage
\section{Derivations}
We provide theoretical derivations of why the multi-view mini-batch strategy is superior to the single-view mini-batch strategy in NeRF and 3DGS in \cref{subsec:derv1}. In addition, we derive the situation where the invalid addition of gradients in different image spaces leads to an unintended consequence in \cref{subsec:derv2}.
\subsection{Minimize variance of mini-batch gradient}
\label{subsec:derv1}
\begin{lemma}
There are $N$ distributions $\mathcal{D}_i(\mu_i, \sigma_i^2), i \in [1, N]$  where $\mathcal{D}_i$ indicates an arbitrary distribution with mean $\mu_i$ and variance $\sigma_i^2$, where $\mu_i \neq \mu_j$ for $i\neq j$. For two cases which are 1) draw $K$ sample from each $\mathcal{D}_i$ and 2) choose $i$ first, then draw $NK$ i.i.d samples from $\mathcal{D}_i$, the former has a smaller variance of the sample mean than the latter.
\end{lemma}
\begin{proof}
Let random variable $X_i \sim \mathcal{D}_i$, and generalize two cases as follows: sample $m$ distributions without replacement, and draw $\frac{NK}{m}$ i.i.d samples from each distribution. For simplicity, assume that $NK$ is divisible by $m$. Let $\Scal$ be indices of selected distributions where $|\Scal|=m$ and $X_{i, k}$ be the $k$-th sample from $\mathcal{D}_{i}$ where $k\in[1, \frac{NK}{m}]$. Then,the  sample mean from the selected distributions $\Scal$  is
\begin{align}
\bar{Z}_{\Scal} = \frac{1}{N}\sum_{j\in\Scal}\sum_{k=1}^{\frac{NK}{m}} X_{j,k}
\end{align}
To calculate the variance of $\bar{Z}$, we can use the law of total variance.
\begin{align}
\mathbb{V}(\bar{Z}) = \mathbb{E}_{\Scal}[\mathbb{V}(\bar{Z}|\Scal)] + \mathbb{V}_{\Scal}(\mathbb{E}[\bar{Z}|\Scal])
\end{align}
\begin{align}
&\mathbb{V}(\bar{Z}|\Scal) = \frac{1}{N^2}\sum_{j\in\Scal}\mathbb{V}(\sum_{k=1}^\frac{NK}{m}X_{{j},k})=\frac{K}{mN}\sum_{j\in\Scal}\sigma_{j}^2 \\
&\mathbb{E}(\bar{Z}|\Scal) = \frac{1}{N}\sum_{j\in\Scal}\mathbb{E}[\sum_{k=1}^{\frac{NK}{m}}X_{{j},k}|\Scal]\\
&=\frac{1}{N}\sum_{j\in\Scal}\mu_{j}\cdot\frac{NK}{m}=\frac{K}{m}\sum_{j\in\Scal}\mu_{j}
\end{align}
Therefore, variance of $\bar{Z}$ is
\begin{align}
    \frac{K}{mN}\mathbb{E}_{\Scal
    }[\sum_{j\in\Scal}\sigma_{j}^2] + \frac{K^2}{m^2}\mathbb{V}_{\Scal}(\sum_{j\in\Scal}\mu_{j})
\end{align}
The first term is constant regardless of $m$ as follows
\begin{align}
    \frac{K}{mN}\mathbb{E}[\sum_{j\in\Scal}\sigma_{j}^2] = \frac{K}{mN} \cdot \frac{{N-1 \choose m-1}}{{N \choose m}}\sum_{i=1}^N\sigma_i^2=\frac{K}{N^2}\sum_{i=1}^N\sigma_i^2
\end{align}
The second term is calculated as follows. \\
For $\mu_{\Scal} = \sum_{j\in\Scal}\mu_{j}$ and $\mu' = \sum_{i=1}^N\mu_i$,
\begin{align}
    &\mathbb{E}_{\Scal}[\mu_{\Scal}] =  \frac{{N-1 \choose m-1}}{{N \choose m}}\mu'=\frac{m}{N}\mu' \\
    &\mathbb{V}_{\Scal}[\mu_{\Scal}] = \mathbb{E}_{\Scal}[(\mu_{\Scal} - \mathbb{E}_{\Scal}[\mu_{\Scal}])^2] \\
\end{align}
Let  $\mathfrak{S}$ is a set of all possible $\Scal$, then 
\begin{align}
    &\mathbb{V}_{\Scal}[\mu_{\Scal}] = \frac{1}{{N \choose m}}\sum_{\Scal \in  \mathfrak{S}}\bigl(\mu_{\Scal}^2 - \frac{2m}{N}\mu_{\Scal}\mu' + \frac{m^2}{N^2}(\mu')^2\bigr)
\end{align}
Let us compute the first term in advance.
\begin{align}
    &\frac{1}{{N \choose m}}\sum_{\Scal \in  \mathfrak{S}}\mu_{\Scal}^2\\
    &=\frac{1}{{N \choose m}}\{{{N-1} \choose {m-1}}\sum_{i=1}^N\mu_i^2 + {{N-2} \choose {m-2}} \sum_{i=1, j\neq i}^N 2\mu_i\mu_j\} \\
    &=\frac{m}{N}(\sum_{i=1}^N\mu_i^2 + \frac{m-1}{N-1}\sum_{i=1, j\neq i}^N 2\mu_i\mu_j)
\end{align}
The second and third terms are calculated as
\begin{align}
    &\frac{1}{{N \choose m}}\sum_{\Scal \in \mathfrak{S}}(-\frac{2m}{N}\mu_{\Scal}\mu' + \frac{m^2}{N^2}(\mu')^2) \\
    &=\frac{1}{{N \choose m}}\{-\frac{2m}{N}\mu'\sum_{\Scal \in  \mathfrak{S}}\mu_{\Scal} + {N \choose m} \frac{m^2}{N^2}(\mu')^2  \}\\
    &=\frac{1}{{N \choose m}}\{-{N-1 \choose m-1}\frac{2m}{N}(\mu')^2 + {N \choose m} \frac{m^2}{N^2}(\mu')^2  \}\\
    &=\frac{(\mu')^2}{{N \choose m}}(\frac{m^2}{N^2} {N \choose m} - \frac{2m}{N}{N-1 \choose m-1}) =-\frac{m^2}{N^2}(\mu')^2 
\end{align}
Using the fact that $(\mu')^2 = \sum_{i=1}^N\mu_i^2+\sum_{i=1, j\neq i}^N 2\mu_i\mu_j$, sum of all three terms is calculated as
\begin{align}
    &\frac{m(N-m)}{N}(\frac{(\mu')^2}{N}-\frac{1}{N-1}\sum_{i=1, j\neq i}^N 2\mu_i\mu_j) \\
    &=\frac{Cm(N-m)}{N}
\end{align}
It can be proved that $C$ is positive as follows 
\begin{align}
    &C = \frac{(\mu')^2}{N}-\frac{1}{N-1}\sum_{i=1, j\neq i}^N 2\mu_i\mu_j\\
    &=\frac{(\mu')^2}{N} - \frac{1}{N-1}\{(\mu')^2-\sum_{i=1}^N\mu_i^2\}
\end{align}
We can utilize $\sigma_\mu^2 = \frac{1}{N}\sum_{i=1}^N\mu_i^2-(\frac{\mu'}{N})^2$.
\begin{align}
    &\frac{(\mu')^2}{N} - \frac{1}{N-1}\{(\mu')^2-\sum_{i=1}^N\mu_i^2\} \\
    &=\frac{(\mu')^2}{N} - \frac{1}{N-1}\{(\mu')^2-(N\sigma_\mu^2+\frac{(\mu')^2}{N})\} \\
    &=\frac{(\mu')^2}{N} - \frac{1}{N-1}\{(1-\frac{1}{N})(\mu')^2-N\sigma_\mu^2\} \\
    &=\frac{N}{N-1}\sigma_\mu^2 > 0
\end{align}
Then, our original goal $\frac{K^2}{m^2}\mathbb{V}_{\Scal}(\sum_{j\in\Scal}\mu_{j})$ is 
\begin{align}
    &\frac{K^2}{m^2}\mathbb{V}_{\Scal}(\sum_{j\in\Scal}\mu_{j})=\frac{K^2}{m^2}\frac{Cm(N-m)}{N}\\
    &=\frac{K^2\sigma_\mu^2}{N-1}(\frac{N}{m}-1)
\end{align}
and $\mathbb{V}(\bar{Z})$ is
\begin{align}
\mathbb{V}(\bar{Z}) = \frac{K}{N^2}\sum_{i=1}^N\sigma_i^2 + \frac{K^2\sigma_\mu^2}{N-1}(\frac{NK}{m}-1) 
\end{align}
Therefore, $\mathbb{V}(\bar{Z}) \propto \frac{1}{m}$. Case 1 and Case 2 correspond to $m=N$ and $m=1$, so the variance of $\bar{Z}$ is larger for Case 2. This lemma also holds for the multivariate case. For $D$-dimensional vectors $\mathbf{x} = (x_1\;x_2 \dots x_D$), $\bmu = (\mu_1 \;\mu_2 \dots \mu_D)$
\begin{align}
    &\mathbb{V}(\mathbf{x}) := \mathbb{E}[||\mathbf{x} - \bmu||^2_2] \\
    &=\mathbb{E}[\sum_{i=1}^D(x_i-\mu_i)^2]=\sum_{i=1}^D \mathbb{E}[(x_i-\mu_i)^2]
\end{align}
Then, we can apply the proof for each dimension, thereby it holds for the multivariate case because if variances of each dimension are larger, so is the sum of them.
\end{proof}

\begin{proposition}
To reduce the variance of the mini-batch gradient, it is more desirable to design a mini-batch to sample data across clusters than to sample all the data from a single cluster.
\end{proposition}
\begin{proof}
For $\psi(\mathbf{x}, y)=\nabla_\mathbf{w} \mathcal{L}(\mathbf{w}, \mathbf{x}, y)$, assume that we have a clustering method to separate a group of $\psi(\mathbf{x}, y)$, and clusters of $\mathbf{x}$ are the same as the clusters of $\psi(\mathbf{x}, y)$. For each cluster $\Ccal_j$, $\psi_j(\mathbf{x}, y) \in \Ccal_j$ follow $\mathcal{D}_{\psi_j}(\bmu_{\psi_j}, \Sigma_{\psi_j})$ for arbitrary distribution $\mathcal{D}_{\psi_j}$. 
For a set $\Scal$ that consists of sampled $\psi(\mathbf{x}, y)$, let $|\Scal|_j$ be the number of samples that satisfy $\psi(\mathbf{x}, y) \in \Ccal_j$. Let $\mathfrak{S_1}$ the set of all possible sets that satisfy $\forall_j|\mathcal{S}|_j=K$ and $\mathfrak{S_2}$ that satisfy $\exists_j \forall_{k\neq j} |\mathcal{S}|_j=NK, |\mathcal{S}|_k=0$. Then by Lemma 1,
\begin{align}
    \label{eq:prop1_eq}
    \mathbb{V}_{\mathfrak{S}_1}(\mathbb{E}_{\psi(\mathbf{x}, y)\in\Scal}[\psi(\mathbf{x}, y)]) < \mathbb{V}_{\mathfrak{S}_2}(\mathbb{E}_{\psi(\mathbf{x}, y)\in\Scal}[\psi(\mathbf{x}, y)])
\end{align}
Because we assume that clusters of $\mathbf{x}$ are the same as the clusters of $\psi(\mathbf{x}, y)$, the proposition holds.

\end{proof}
\begin{lemma}
    For a single ray  $\mathbf{r}(t)=\mathbf{o}+t\mathbf{d}$, the intersection point on the ray $\mathbf{r}(t_i)$ and weights of NeRF $\mathbf{w_\sigma}$ and $\mathbf{w_{\mathbf{c}}}$, update of NeRF is mostly contributed by the intersection point of the surface on the ray. In other words,  $\frac{\pr \cl}{\pr \cw} \propto \frac{\pr \mathbf{c}_{\mathbf{w_{\mathbf{c}}}}(t_i)}{\pr \cw}$ and $\frac{\pr \cl}{\pr \dw} \propto \frac{\pr \sigma_{\mathbf{w_\sigma}}(t_i)}{\pr \dw}$.
\end{lemma}
\begin{proof}
    For a ray $\mathbf{r}(t)=\mathbf{o}+t\mathbf{d}$, the color of the ray is rendered through volumetric rendering as follows
\begin{align}
    &\hat{\mathbf{C}}(\mathbf{r}) = \int_{t_n}^{t_f}T_{\dw}(t)\sigma_{\dw}(\mathbf{r}(t))\mathbf{c}_{\cw}(\mathbf{r}(t), \mathbf{d})dt \\
    &T_{\dw}(t) = \exp(-\int_{t_n}^t \sigma_{\dw}(\mathbf{r}(s))ds)
\end{align}
where $\mathbf{w_\sigma}$ and $\mathbf{w_{\mathbf{c}}}$ are weights of a neural network that predicts color and density. For simplicity, we omit $\mathbf{d}, \mathbf{r}$, and replace the following statements.
\begin{align}
\mathbf{r}(t) \Rightarrow t \quad T_{\dw} \Rightarrow T \quad  \sigma_{\dw} \Rightarrow \sigma \quad \mathbf{c}_{\cw} \Rightarrow \mathbf{c}
\end{align} 
Then, $\frac{\pr \cl}{\pr \cw}$ and $\frac{\pr \cl}{\pr \mathbf{w}_{\sigma}}$ are as follows.
\begin{align}
&\frac{\pr \cl}{\pr \cw} = \frac{\pr \cl}{\pr \hat{\mathbf{C}}} \frac{\pr \hat{\mathbf{C}}}{\pr \cw} \\
&= \frac{\pr \cl}{\pr \hat{\mathbf{C}}}\int_{t_n}^{t_f}T(t)\sigma(t)\frac{\pr \mathbf{c}(t)}{\pr \cw}dt \\
&\frac{\pr \cl}{\pr \dw} = \frac{\pr \cl}{\pr \hat{\mathbf{C}}} \frac{\pr \hat{\mathbf{C}}}{\pr \dw} \\
&= \frac{\pr \cl}{\pr \hat{\mathbf{C}}}\int_{t_n}^{t_f}\frac{\pr T(t)}{\pr \dw}\sigma(t)\mathbf{c}(t) + T(t)\frac{\pr \sigma (t)}{\pr \dw}\mathbf{c}(t)dt \\
&\text{where } \frac{\pr T(t)}{\pr \dw }=-T(t)\int_{t_n}^t \frac{\pr \sigma(s)}{\pr \dw}ds
\end{align}
For a 3D point $\mathbf{r}(t_i)$ to be an intersection point of the surface, $\sigma(j) \approx 0$ for $j < i$, $T(t_i) \approx 1$ and $T(k) \approx 0$ (collorary, $\frac{\pr T(t)}{\pr \dw } \approx \mathbf{0}$)  for $k > i$. We can regard $\frac{\pr \sigma(j)}{\pr \dw } \approx \mathbf{0}$ because NeRF uses ReLU/Softplus functions as an activation function for $\sigma$.

Then, $T(t)\sigma(t)$ and $T(t)\frac{\pr \sigma(t)}{\pr \dw }$ can be considered direc-delta function. Then, gradient of $\cw$ is,
\begin{align}
&\int_{t_n}^{t_f}T(t)\sigma(t)\frac{\pr \mathbf{c}(t)}{\pr \cw}dt \\
& = \int_{t_n}^{t_f}\frac{\pr \mathbf{c}(t)}{\pr \cw}\delta(t-t_i) dt=\frac{\pr \mathbf{c}(t_i)}{\pr \cw} 
\end{align}
Similarily, gradient of $\dw$ is,
\begin{align}
&\frac{\pr \cl}{\pr \mathbf{w}_{\sigma}}=\int_{t_n}^{t_f} T(t) \frac{\pr \sigma(t)}{\pr \dw}\mathbf{c}(t)dt+\int_{t_n}^{t_f} \frac{\pr T(t)}{\pr \dw} \sigma(t)\mathbf{c}(t)dt\\
&=\int_{t_n}^{t_f} T(t_i)\mathbf{c}(t)\frac{\pr \sigma(t_i)}{\pr \dw}\delta(t-t_i)dt \\
&=T(t_i)c(i)\frac{\pr \sigma(t_i)}{\pr \dw} \\
&\int_{t_n}^{t_f} \frac{\pr T(t)}{\pr \dw} \sigma(t)\mathbf{c}(t)dt \\
&=\int_{t_n}^{t_f} -T(t)(\int_{t_n}^t \frac{\pr \sigma(s)}{\pr \dw}ds) \sigma(t)\mathbf{c}(t)dt \\
&=\int_{t_n}^{t_f} -T(t_i)(\int_{t_n}^t \frac{\pr \sigma(s)}{\pr \dw}ds) \sigma(t_i)\mathbf{c}(t)\delta(t-t_i)dt \\
&=-T(t_i)(\int_{t_n}^{t_i} \frac{\pr \sigma(s)}{\pr \dw}ds) \sigma(t_i)\mathbf{c}(t_i) \\
&=-T(t_i)(\int_{t_n}^{t_i} \frac{\pr \sigma(t_i)}{\pr \dw}\delta(s-t_i)ds) \sigma(t_i)\mathbf{c}(t_i)\\
&=-T(t_i)\frac{\pr \sigma(t_i)}{\pr \dw}\sigma(t_i)\mathbf{c}(t_i)
\end{align}
Therefore, it is proven that $\frac{\pr \cl}{\pr \cw} \propto \frac{\pr \mathbf{c}_{\mathbf{w_{\mathbf{c}}}}(t_i)}{\pr \cw}$ and $\frac{\pr \cl}{\pr \dw} \propto \frac{\pr \sigma_{\mathbf{w_\sigma}}(t_i)}{\pr \dw}$.

\begin{figure}[t]
     \centering
     \includegraphics[trim=0 0 0 0,clip, width=0.7\linewidth]{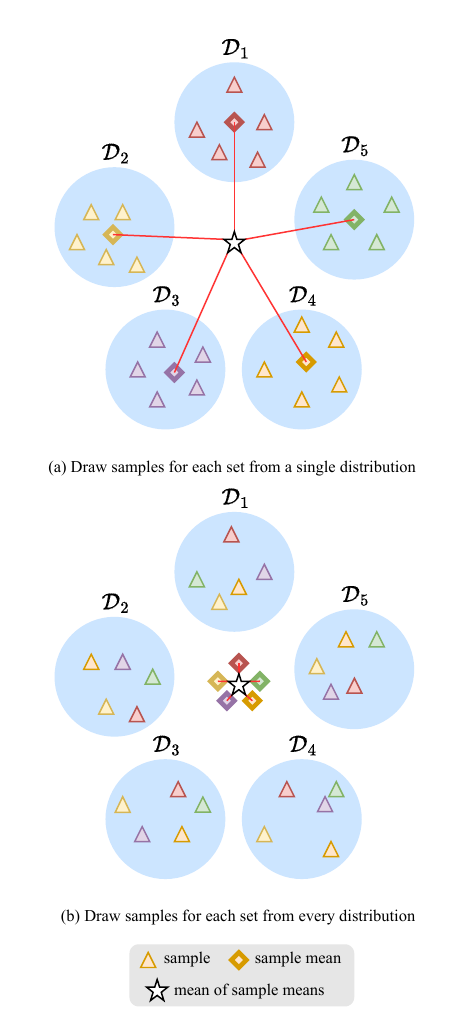}
     \vspace{-5pt}
     \caption{\textbf{Illustration of the two cases in proposition 1}. The figure demonstrates the situation where we draw five samples for each set, composing five sets. Triangles with the same colors indicate samples in the same set. As shown in (a), drawing samples from a single distribution for each set causes each sample mean to be dispersed, increasing the variance of the sample mean. In contrast, (b) suggests drawing a sample from every distribution makes the sample means similar, reducing the variance of the sample mean.}
     \label{fig:sup_gradient}
\end{figure}

\noindent We provide an illustration of the two cases in \cref{fig:sup_gradient}.

\end{proof}
\begin{proposition}
It is more desirable in terms of NeRF optimization to sample rays from individual pixels than patches, and across multiple images than from a single image.
\end{proposition}
\begin{proof}
By Lemma 2, the dataset of NeRF $\{\mathbf{r}_j(t), \mathbf{C}(\mathbf{r}_j)\}$ is equivalent to $\{\mathbf{r}_j(t_i), \mathbf{C}(\mathbf{r}_j)\} \text{ where } \mathbf{r}_j(t_i)$ is an intersection point on $\mathbf{r}_j(t)$. By Proposition 1, we can infer that mini-batch in NeRF should avoid having ray pairs that are $\mathbf{r}_{j_1}(t_i) \approx \mathbf{r}_{j_2}(t_i)$ to reduce the variance of mini-batch. Then, it is evident that rays across multiple images are less likely to satisfy such a condition than ones in a single image because of the locality of pixel dependencies. 
\end{proof}

\subsection{Adding 2D gradients in different spaces}
\label{subsec:derv2}
For perspective projection matrix $P$, extrinsic matrix $[R|t]$, world space point $\bmu$, camera space point $\bmu_{\text{cam}}$, and NDC space point $\bmu_{\text{NDC}}$, they have following relationship.
\begin{align}
&\bmu=(x,y,z)\\
&\bmu_{\text{cam}}=(x_{\text{cam}}, y_{\text{cam}}, z_{\text{cam}})\\
&\bmu_{\text{NDC}}=(x_{\text{NDC}}, y_{\text{NDC}}, z_{\text{NDC}}) 
\end{align}
\begin{align}
&(\bmu,1)[R|t]^T=(\bmu,1)
\begin{pmatrix}
R_0&R_1&R_2&0\\
R_3&R_4&R_5&0\\
R_6&R_7&R_8&0\\ 
t_0&t_1&t_2&1
\end{pmatrix}\\
&=(\bmu_{\text{cam}},1) \\
&(\bmu_{\text{cam}},1)P^T = (\bmu_{\text{cam}},1)
\begin{pmatrix}
P_0&0&0&0\\
0&P_1&0&0\\
0&0&P_2&1\\
0&0&P_3&0
\end{pmatrix}\\
&=
(\hat{\bmu}_{\text{NDC}},w_{\text{NDC}}) \\
&\qquad\qquad\qquad\bmu_{\text{NDC}}=\frac{1}{w_{\text{NDC}}}\hat{\bmu}_{\text{NDC}}
\end{align}
\begin{align}
(\frac{d\cl}{d\bmu_{\text{cam}}})^T = (\frac{d\cl}{d\bmu_{\text{NDC}}})^T 
\begin{pmatrix}
\frac{P_0}{z_{\text{cam}}}&0&-\frac{x_{\text{NDC}}}{P_0}\\
0&\frac{P_1}{z_{\text{cam}}}&-\frac{y_{\text{NDC}}}{P_1}\\
0&0&0\\
\end{pmatrix}
\\
(\frac{d\cl}{d\bmu})^T = (\frac{d\cl}{d\bmu_{\text{NDC}}})^T 
\begin{pmatrix}
\frac{P_0}{z_{\text{cam}}}&0&-\frac{x_{\text{NDC}}}{P_0}\\
0&\frac{P_1}{z_{\text{cam}}}&-\frac{y_{\text{NDC}}}{P_1}\\
0&0&0\\
\end{pmatrix}
R
\end{align}
Let us think of a situation where two viewpoints are located exactly opposite each other with respect to the origin with both facing the origin. $R_1 = \text{diag}(1, 1, 1), t_1 = (0, 0, -1), R_1 = \text{diag}(-1, 1, -1), t_1 = (0, 0, 1)$, and 3D positional gradient $\frac{d\cl}{d\bmu} = (1,0,0).$ Then,
\begin{align}
&(\frac{d\cl}{dx_{\text{NDC}}^{(1)}}, \frac{d\cl}{dy_{\text{NDC}}^{(1)}}) = (\frac{d\cl}{dx}, \frac{d\cl}{dy})
\begin{pmatrix}
\frac{z_{\text{cam}}}{P_0}&0\\
0&\frac{z_{\text{cam}}}{P_1}\\
\end{pmatrix} \\
&(\frac{d\cl}{dx_{\text{NDC}}^{(2)}}, \frac{d\cl}{dy_{\text{NDC}}^{(2)}}) = (\frac{d\cl}{dx}, \frac{d\cl}{dy})
\begin{pmatrix}
-\frac{z_{\text{cam}}}{P_0}&0\\
0&\frac{z_{\text{cam}}}{P_1}\\
\end{pmatrix}
\end{align}
Because $\frac{dL}{dy} = 0$, the sum of two vectors is nullified even for the same 3D positional gradient.
\section{Additional experiment}
We compare different numbers of viewpoints in each iteration in \cref{tab:batch_size}. We observe that performance increases if we add the number of viewpoints for a mini-batch, but it should be noticed that more viewpoints in a minibatch is equivalent to more memory consumption and time overhead.
\begin{table}
    \begin{minipage}{1.0\linewidth}
    \centering\setlength\tabcolsep{2.3pt}
        \vspace{1mm}
        \centering\resizebox{0.9\linewidth}{!}{
            \begin{tabular}{lcccccccc}
            \toprule
                & \multicolumn{4}{c}{\small3DGS} & \multicolumn{4}{c}{\small3DGS-MCMC} \\
                \cmidrule(lr){2-5} \cmidrule(lr){6-9}
                 & base & 2 & 4 & 8 & base & 2 & 4 & 8 \\
                 \midrule
                 PSNR $\uparrow$       & 29.28 & 29.47 & 29.74 & \textbf{29.94} & 29.83 & 30.19 & 30.42 & \textbf{30.56} \\
                 SSIM $\uparrow$       & 0.878 & 0.883 & 0.886 & \textbf{0.887} & 0.895 & 0.900 & \textbf{0.901} & \textbf{0.901} \\
                 LPIPS $\downarrow$    & 0.167 & 0.156 & 0.154 & \textbf{0.154} & 0.145 & 0.140 & \textbf{0.138} & \textbf{0.138} \\
                 \bottomrule
            \end{tabular}
    }
    \caption{\textbf{Comparison by number of viewpoints in each iteration.} Results are the average values on the MipNeRF-360 dataset.}
    \label{tab:batch_size}
    \end{minipage}
\end{table}

\section{Experiment details}
When applying multi-view training, we found that the learning pattern changed, so in 3DGS, the threshold hyperparameter for pruning Gaussians was multiplied by the number of images, and in 3DGS-MCMC, the mean noise learning rate was divided by the number of images. Other hyperparameters are the same as before.
\section{Rasterization in multi-view training}
We describe the modifications made to the 3DGS rasterizer for multi-view training. The 3DGS rasterizer primarily consists of two main components: preprocessing and rendering.  In the preprocessing stage, the rasterizer performs the following steps:  
\begin{enumerate}  
    \item Frustum culling  
    \item Computing projected 2D Gaussians  
    \item Duplicating projected Gaussians for each tile they cover  
    \item Sorting Gaussians by depth for each tile  
\end{enumerate}  
For multi-view training, we introduce the following modifications. First, to reduce memory consumption, the rasterizer determines which Gaussians participate in rendering for each viewpoint in a mini-batch before preprocessing. Then, memory is allocated only for these participating Gaussians. Since only a subset of Gaussians is used in rendering, this approach significantly reduces memory usage. (For $B$ viewpoints in a mini-batch, with $P$ Gaussians in total, and $N$ Gaussians remaining after this step, typically $N \ll B \times P$.) After memory allocation, the four preprocessing steps are executed.  It is important to note that after projection, a single Gaussian can appear multiple times in the array of projected Gaussians if it is present in multiple viewpoints. To account for this, we modify the sorting step to ensure that projected Gaussians from different viewpoints are handled separately. In the original 3DGS rasterizer, after duplicating projected Gaussians, each duplicated Gaussian is assigned a unique 64-bit ID, composed of a 32-bit tile index and a 32-bit depth value. This ensures that projected Gaussians in each tile are automatically separated during sorting.  Similarly, we assign each Gaussian an ID consisting of a 16-bit tile index, a 16-bit viewpoint index (indicating the viewpoint where the Gaussian is projected), and a 32-bit depth value. This guarantees that projected Gaussians are separated by both tile and viewpoint. The rasterizer then allocates multiple blocks per tile—one block for each viewpoint—and each block renders its designated pixels using the projected Gaussians corresponding to its assigned viewpoint.  

\section{Limitations}
Even though we successfully mitigate the overhead of rendering part in rasterization, the overhead caused by the preprocessing step still remains. It is necessary to eliminate the overhead of the preprocessing step to achieve constant time complexity concerning the number of images as NeRF. In addition, we observe that multi-view mini-batch training causes overfitting for a few scenes, including \textit{bicycle/stump} scenes in the MipNeRF-360 dataset and scenes in the Deep-Blending dataset. We leave those as intriguing future work.
\section{Detailed results}
We provided full results of the main table in \cref{table:sup_full_table}.
\begin{table*}[t]
    \centering
    \resizebox{\linewidth}{!}{
    \setlength{\tabcolsep}{4pt}
    \begin{tabular}{@{}c l cc|cc@{}} 
    \toprule
    && 3DGS & 3DGS + Ours & MCMC & MCMC + Ours \\
    && \tiny{PSNR$\uparrow$ / SSIM$\uparrow$ / LPIPS$\downarrow$} & \tiny{PSNR$\uparrow$ / SSIM$\uparrow$ / LPIPS$\downarrow$} & \tiny{PSNR$\uparrow$ / SSIM$\uparrow$ / LPIPS$\downarrow$} & \tiny{PSNR$\uparrow$ / SSIM$\uparrow$ / LPIPS$\downarrow$}   \\
    \midrule 
    \multirow{8}{*}{\rotatebox[origin=c]{90}{MipNeRF-360}} 
    & Bicycle & 25.62 / 0.775 / 0.204 & 25.84 / 0.796 / 0.169 & 26.12 / 0.809 / 0.161 & 26.27 / 0.816 / 0.156 \\
    & Bonsai  & 32.32 / 0.947 / 0.173 & 32.90 / 0.952 / 0.157 & 32.81 / 0.953 / 0.160 & 33.85 / 0.958 / 0.152 \\
    & Counter & 29.11 / 0.915 / 0.179 & 29.63 / 0.920 / 0.167 & 29.45 / 0.924 / 0.163 & 30.09 / 0.930 / 0.152 \\
    & Garden  & 27.75 / 0.872 / 0.102 & 28.22 / 0.879 / 0.102 & 28.12 / 0.883 / 0.090 & 28.70 / 0.893 / 0.081 \\
    & Kitchen & 31.57 / 0.932 / 0.113 & 32.49 / 0.938 / 0.106 & 32.24 / 0.938 / 0.108 & 33.24 / 0.945 / 0.099 \\
    & Room    & 31.67 / 0.927 / 0.191 & 32.44 / 0.932 / 0.179 & 32.38 / 0.937 / 0.171 & 33.27 / 0.942 / 0.163 \\
    & Stump   & 26.88 / 0.780 / 0.209 & 26.70 / 0.785 / 0.196 & 27.72 / 0.820 / 0.163 & 27.50 / 0.819 / 0.163 \\
    \cmidrule(lr{0.5em}){2-6}
    & Average & 29.28 / 0.878 / 0.167 & 29.74 / 0.886 / 0.154 & 29.83 / 0.895 / 0.145 & 30.42 / 0.901 / 0.138 \\
    \midrule 
    \multirow{3}{*}{\rotatebox[origin=c]{90}{\tiny\makecell{Tank~\&\\Temples}}} 
    & Train & 21.96 / 0.810 / 0.207 & 22.20 / 0.824 / 0.189 & 22.71 / 0.840 / 0.181 & 23.28 / 0.850 / 0.172 \\
    & Truck & 25.43 / 0.879 / 0.146 & 25.85 / 0.889 / 0.128 & 26.46 / 0.898 / 0.109 & 26.85 / 0.901 / 0.117 \\
    \cmidrule(lr{0.5em}){2-6}
    & Average & 23.69 / 0.845 / 0.177 & 24.02 / 0.856 / 0.158 & 24.59 / 0.869 / 0.145 & 25.06 / 0.876 / 0.144 \\
    \midrule 
    \multirow{3}{*}{\rotatebox[origin=c]{90}{\tiny\makecell{Deep\\Blending}}} 
    & Dr Johnson & 29.07 / 0.899 / 0.244 & 29.40 / 0.901 / 0.238 & 29.01 / 0.900 / 0.241 & 29.89 / 0.912 / 0.221 \\
    & Playroom & 30.03 / 0.906 / 0.243 & 29.87 / 0.904 / 0.238 & 30.13 / 0.910 / 0.239 & 30.34 / 0.916 / 0.227 \\
    \cmidrule(lr{0.5em}){2-6}
     & Average & 29.55 / 0.902 / 0.244 & 29.64 / 0.903 / 0.238 & 29.57 / 0.905 / 0.240 & 30.11 / 0.914 / 0.224 \\
     \midrule 
     \multirow{9}{*}{\rotatebox[origin=c]{90}{OMMO}} 
    & 01 & 25.66 / 0.787 / 0.213 & 25.93 / 0.803 / 0.196 & 25.81 / 0.800 / 0.189 & 26.26 / 0.812 / 0.176 \\
    & 03 & 26.10 / 0.867 / 0.212 & 27.30 / 0.890 / 0.182 & 27.61 / 0.895 / 0.177 & 28.55 / 0.910 / 0.159 \\
    & 05 & 28.75 / 0.882 / 0.224 & 29.01 / 0.887 / 0.191 & 29.28 / 0.889 / 0.211 & 29.84 / 0.896 / 0.201 \\
    & 06 & 26.82 / 0.919 / 0.179 & 27.36 / 0.925 / 0.162 & 27.77 / 0.948 / 0.125 & 28.94 / 0.952 / 0.120 \\
    & 10 & 30.10 / 0.903 / 0.171 & 31.45 / 0.919 / 0.145 & 31.86 / 0.920 / 0.142 & 33.23 / 0.932 / 0.123 \\
    & 13 & 33.61 / 0.957 / 0.110 & 35.23 / 0.966 / 0.094 & 33.59 / 0.957 / 0.103 & 35.56 / 0.968 / 0.081 \\
    & 14 & 31.80 / 0.951 / 0.093 & 32.28 / 0.956 / 0.085 & 31.78 / 0.954 / 0.089 & 32.64 / 0.960 / 0.082 \\
    & 15 & 31.08 / 0.944 / 0.091 & 31.52 / 0.945 / 0.092 & 30.78 / 0.947 / 0.087 & 31.76 / 0.953 / 0.080 \\
    \cmidrule(lr{0.5em}){2-6}
    & Average & 29.24 / 0.901 / 0.162 & 30.01 / 0.911 / 0.143 & 29.81 / 0.914 / 0.140 & 30.85 / 0.923 / 0.128 \\
    \bottomrule
    \end{tabular}
    }
    \caption{
    We report the full table of 3DGS~\citep{kerbl3Dgaussians}, 3DGS-MCMC~\citep{kheradmand20243d} and the results of adding our method to them. MCMC is an abbreviation of 3DGS-MCMC~\citep{kheradmand20243d}. We report the average values of three repeated experiments.
    }
    \label{table:sup_full_table}
\end{table*}

\section{Algorithms}
\label{sup:algorithms}
We provide detailed pseudocodes of the difference between the mini-batch strategy of NeRF and 3DGS in \cref{alg:nerf_vs_gs}, tiled rendering of 3DGS in \cref{alg:tile_rendering} and thread-efficient partial rendering in \cref{alg:agg_rendering}.
\begin{algorithm}
\label{alg:rendering}
\caption{Mini-batch strategy of NeRF and 3DGS 
}
\begin{algorithmic}
\State \textit{// Multi-view mini-batch of NeRF }
\For{$i$ = 1 \dots $N$}
    \State $\{\mathbf{r}_{ij}\}_{j=1}^M = \textit{generateRay}(\{p_{ij}\}_{j=1}^M, P_i)$    
\EndFor
\State $\mathcal{D} = \{(\mathbf{r}_{q}, \mathbf{C}_{q})\}_{q=1}^{MN} = \textit{concat}(\{(\mathbf{r}_{ij}, \mathbf{C}_{ij})\}_{j=1}^M, \textit{dim=i})$
\State $\mathcal{B}_{\text{NeRF}} = \textit{sample}(\mathcal{D}, \textit{count=}B)$\\
\State \textit{// Single-view mini-batch of 3DGS }
\For{$i$ = 1 \dots $N$}
    \State $\mathcal{D}_i = \{(p_{ij}, \mathbf{C}_{ij})\}_{j=1}^M$
\EndFor
\State $\mathcal{D} = \{\mathcal{D}_i\}_{i=1}^N$
\State $\mathcal{B}_{\text{3DGS}} = \textit{sample}(\mathcal{D}, \textit{count=}1)$

\end{algorithmic}
\label{alg:nerf_vs_gs}
\end{algorithm}
\begin{algorithm}
\label{alg:rendering}
\caption{Tiled rendering of 3DGS \\
$i$ : index of a thread in a tile \\
$D$ : sorted array of projected Gaussian primitives for a tile \\
$P_{i}$ : image pixel  \\
$N$ : The number of pixels (= threads) in a tile \\
$V_i$ : temporal variable for saving intermediate values of alpha blending
}
\begin{algorithmic}
\State $\text{offset} = 0$ 
\State $\text{Shmem}[N]\text{ // shared memory across }i$
\For{\text{thread }$i$ = 1 \dots $N$ in a block}
\State $\text{isValid}_i$ = $\text{true}$
\If{$P_i$ is outside the image}
\State $\text{isValid}_i \gets \text{false}$
\EndIf
\While{unfetched data left in $D$}
\State $\text{Shmem}[i] \gets D[\text{offset} \cdot N + i]$
\If{$\text{isValid}_i$}
\State $V_i \gets \textit{alphaBlending}(\text{Shmem}, V_i)$
\EndIf
\State \textit{synchronize}() \text{// wait for other threads}
\State offset $\gets$ offset + 1
\EndWhile
\State $P_i \gets V_i$
\EndFor
\end{algorithmic}
\label{alg:tile_rendering}
\end{algorithm}
\begin{algorithm}
\label{alg:rendering}
\caption{Thread-efficient partial rendering \\
$i$ : index of a thread in a tile \\
$I$ : a viewpoint \\
$D_{I}$ : sorted array of Gaussian primitives that are projected on image plane of $I$. \\
$K$ : the number of participating viewpoints \\
$P$ : image pixel  \\
$N$ : the number of pixels (= threads) in a tile \\
$A$ : an array that maps a thread to the pixel index to render \\
$V_{k,i}$ : temporal variable for saving intermediate values of alpha blending 
}
\begin{algorithmic}
\State $\text{offset} = 0$
\For{block $k$ = 1 \dots $K$}
\State $\text{Shmem}_k [\frac{N}{K}]\text{ // shared memory across }i$
\For{thread $i$ = $1$ \dots $\frac{N}{K}$ in block $k$}
\State $\text{isValid}_{k,i}$ = $\text{true}$
\If{$P_{A[k,i]}$ is outside the image}
\State $\text{isValid}_{k,i} \gets \text{false}$
\EndIf
\While{unfetched data left in $D_{I_k}$}
\State $\text{Shmem}_k [i] \gets D_{I_k}[\text{offset} \cdot \frac{N}{K} + i]$
\If{$\text{isValid}_{k,i}$}
\State $V_{k,i} \gets \textit{alphaBlending}(\text{Shmem[k]}, V_{k,i})$
\EndIf
\State \textit{synchronize}() \text{// wait for other threads}
\State offset $\gets$ offset + 1
\EndWhile
\State $P_{A[k,i]} \gets V_{k,i}$
\EndFor
\EndFor
\end{algorithmic}
\label{alg:agg_rendering}
\end{algorithm}
\end{document}